\documentclass[12pt]{article}

\usepackage{scicite}

\usepackage{times}

\usepackage{amsmath}
\usepackage{amsfonts}
\usepackage{amssymb}
\usepackage{graphicx}
\usepackage{comment}
\usepackage{subfigure}
\usepackage{url} 
\usepackage{hyperref}
\usepackage{booktabs, makecell, multirow, tabularx}

\newenvironment{sciabstract}{%
\begin{quote} \bf}
{\end{quote}}

\title{Deep learning powered real-time identification of insects using citizen science data}

\author
{Shivani Chiranjeevi,${}^{1}$ Mojdeh Sadaati,${}^{1}$ Zi K Deng,${}^{3}$ Jayanth Koushik,${}^{2}$ \\ Talukder Z Jubery,${}^{1}$  Daren Mueller,${}^{1}$ Matthew E O'Neal,${}^{1}$  Nirav Merchant,${}^{3}$ \\ Aarti Singh,${}^{2}$ Asheesh K Singh,${}^{1}$  Soumik Sarkar,${}^{1}$ \\ Arti Singh,${}^{1\ast}$ Baskar Ganapathysubramanian,${}^{1\ast}$\\
\\
\normalsize{${}^{1}$Iowa State University, IA, USA}\\
\normalsize{Carnegie Mellon University, PA, USA}\\
\normalsize{${}^{2}$University of Arizona, AZ, USA}\\
\\
\normalsize{$^\ast$To whom correspondence should be addressed; E-mail: baskarg@iastate.edu, arti@iastate.edu.}
}

\date{}

\begin{document} 


\baselineskip24pt


\maketitle



\begin{sciabstract}
Insect-pests significantly impact global agricultural productivity and quality. Effective management involves identifying the full insect community, including beneficial insects and harmful pests, to develop and implement integrated pest management strategies. Automated identification of insects under real-world conditions presents several challenges, including differentiating similar-looking species, intra-species dissimilarity and inter-species similarity, several life cycle stages, camouflage, diverse imaging conditions, and variability in insect orientation. A deep-learning model, InsectNet, is proposed to address these challenges. InsectNet is endowed with five key features: (a) utilization of a large dataset of insect images collected through citizen science; (b) label-free self-supervised learning for large models; (c) improving prediction accuracy for species with a small sample size; (d) enhancing model trustworthiness; and (e) democratizing access through streamlined MLOps. This approach allows accurate identification ($>$96\% accuracy) of over 2500 agriculturally and ecologically relevant insect species, including pollinator (e.g., butterflies, bees), parasitoid (e.g., some wasps and flies), predator species (e.g., lady beetles, mantises, dragonflies) and harmful pest species (e.g., armyworms, cutworms, grasshoppers, stink bugs). The model and associated workflows are available through a web-based portal and an easily reusable software stack. InsectNet can identify invasive species, provide fine-grained insect species identification, and work effectively in challenging backgrounds. It also can abstain from making predictions when uncertain, facilitating seamless human intervention and making it a practical and trustworthy tool. InsectNet can guide citizen science data collection, especially for invasive species where early detection is crucial. Similar approaches may transform other agricultural challenges like disease detection and underscore the importance of data collection, particularly through citizen science efforts.
\end{sciabstract}


\section{Introduction}
In the U.S., agriculture, food, and other related industries contributed \$1.26 trillion to the U.S. gross domestic product (GDP) in 2021~\cite{USDA}. Insect pests, observed at all stages of plant growth, negatively affect the quality and quantity of crop yields in agriculture. Accurate detection of insects is imperative for prompt, timely, and optimal decision-making~\cite{hoye2021deep}. Accurate detection allows farmers to identify the specific pest species that cause damage, enabling them to use targeted pest control methods instead of blanket applications of pesticides. This reduces the risk of harm to beneficial insects and other non-target organisms. Furthermore, such an accurate spatio-temporal identification of insects and pests can result in effective pest control measures, which reduce crop losses, increase farm operations' profitability and sustainability, and reduce chemical runoff into water bodies~\cite{water_bodies}. 

Automated approaches for insect detection are becoming increasingly necessary for several reasons. First, manual scouting for identification and quantification of insects and pests is challenging at all farming scales due to the limited availability of experts (especially in remote and rural locations) and expertise levels for accurate identification. Second, rising temperatures are expected to increase the risk of invasion by new pests and transmission of insect-induced diseases~\cite{climate}. Third, several insect-pest species have high fecundity and over-wintering ability -- for example, \textit{Lycorma delicatula} (Spotted Lanternfly (SLF)) -- and consequently exhibit rapid spread across large areas in a limited amount of time, devastating crops, orchards, and logging industries. Fourth, increased trade and travel makes it easier for invasive insect species to access new geographic locations. For example, invasive insect species like SLF have reached several states in the Northeastern and Mid-Atlantic regions of the U.S. threatening crop species ranging from ornamental crops to fruit and tree species~\cite{lanternfly}. SLF is projected to reach and establish in California by 2033 if preventative measures are not taken immediately to limit its spread~\cite{california}.

\begin{figure*}[ht!]
\centering
  \includegraphics[ width=16cm]
  {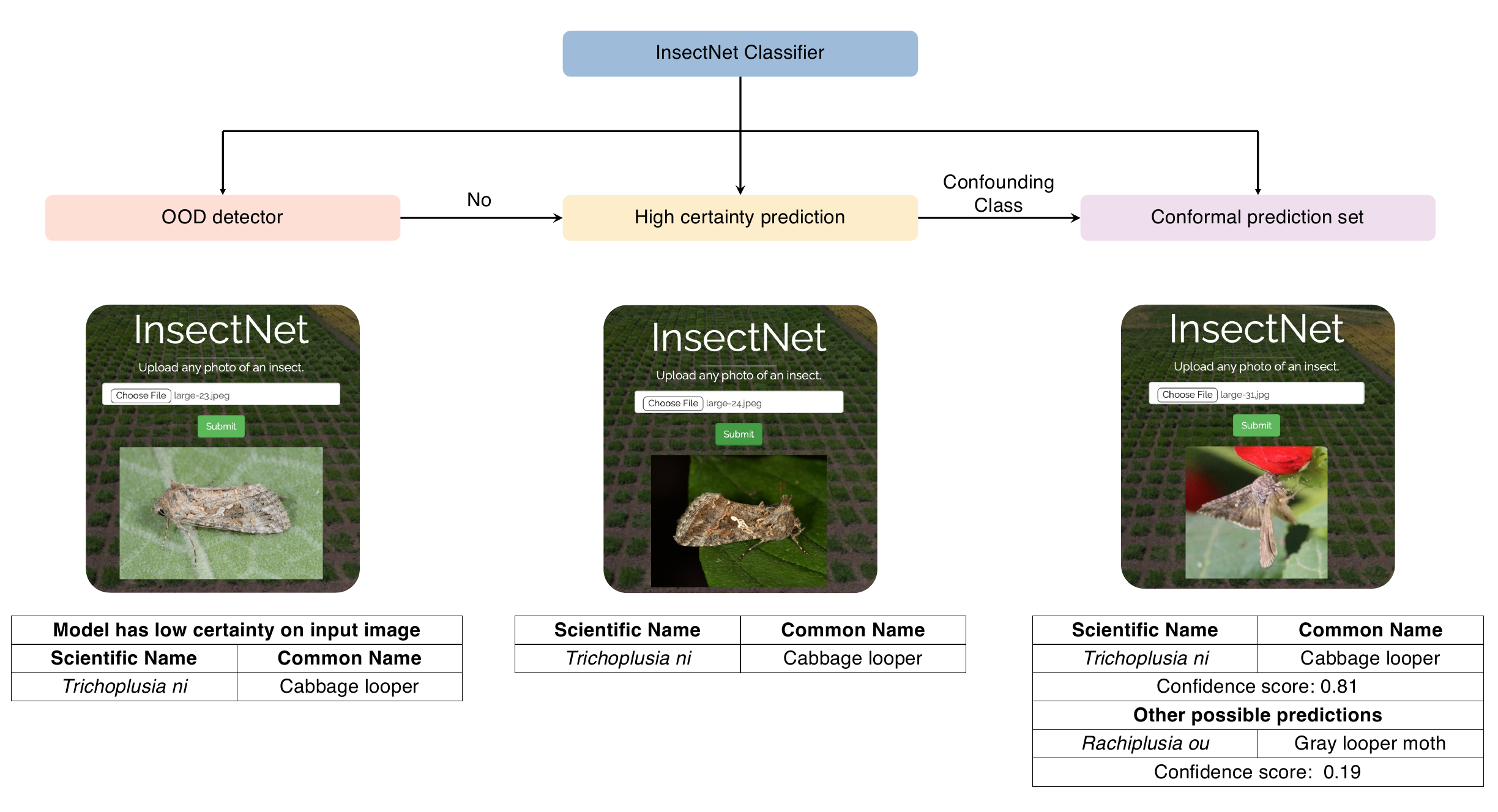}
  \caption{
InsectNet in action. After an image is uploaded, InsectNet first performs out-of-distribution (OOD) detection. (Left) If OOD detection is true, InsectNet provides a warning along with its prediction. (Middle) If not OOD, InsectNet produces a prediction with no warning. (Right) Additionally, InsectNet provides conformal sets with a predefined (here, 97.5\%) confidence. In this instance, the images above all belong to insect species \textit{Trichoplusia ni} (Cabbage looper). The figure on the right is sufficiently confusing for InsectNet to predict a conformal set of two closely related species.}
  \label{Fig:Workflow}
\end{figure*}

The past few years have seen various attempts to automate insects identification, with the earliest attempts using classical ML methods~\cite{svm_ann,mkl}, to the more recent efforts using deep learning-based approaches~\cite{cnn_1,cnn_2,cnn_3,cnn_4,cnn_5,spiesman2021assessing}. Most efforts have focused on utilizing relatively small labeled datasets ($\leq 150,000$ images) spanning a modest number ($<50$) of clearly distinguishable insect species.\footnote{Recent work that uses DNA along with images produces higher accuracy predictions~\cite{badirli2021classifying}, but is rather limited in applicability outside a lab} A comprehensive review uncovered numerous obstacles and shortcomings in the realm of image-based insect detection and classification \cite{amarthunga}. These issues include the narrow scope of current datasets (which only cover a few insect species, natural habitats, and regions), unbalanced datasets that complicate machine learning, unidentified insect species within geographic regions, and difficult scenarios (such as overlapping insects, morphologically similar species, and intra-species variations that complicate image-based identification). Additionally, capturing images of insects throughout their life stages and dealing with fast-moving insects (where blurred images and external factors like lighting conditions may impact accuracy) present challenges. The focus on species-level classification often overlooks valuable information like higher-level taxonomy, sex, and life stage. Lastly, the significant data requirements for deep learning models further complicate the field.
We identify the following challenges that any automated insect identification system must resolve to be useful:
\begin{enumerate}
\setlength{\parskip}{0pt}
\setlength{\itemsep}{0pt plus 1pt}
    \item \textit{Large number of insect species:} Insects constitute the most varied group of species among eukaryotes on earth. We consider over 2500 agriculturally and ecologically relevant insect species, and an automated system should ideally be able to identify across this large number of insect species.
    \item \textit{Metamorphosis (Multiple life cycle stages)} of an insect, where the physical features of each stage in an insect species are wildly dissimilar across its life cycle stages, for e.g., egg, larva, pupa, nymph, and adult.
     \item \textit{Intra-species dissimilarity} due to color and pattern variations within the same species, for example, \textit {Harmonia axyridis} (Asian lady beetle).
    \item  \textit{Inter-species similarity:} Fine-grained classification is needed for several insects belonging to distinct species that are visually very similar. This similarity of features often confuses humans (and human experts), complicating accurate identification. An automated classifier should be able to account for these inter-species similarities during prediction, for example, insect-pest species \textit{Euschistus servus} (Brown Stink Bug) and \textit{Halyomorpha halys} (Brown Marmorated Stink Bug).
   \item \textit{Insect camouflaging and diverse backgrounds:} Insects camouflage with the background, a survival mechanism against predators, which can make automated identification challenging. Diverse backgrounds over which an insect is imaged, the usually small foreground (insect size), and variability in illumination in the field of view; all make identification challenging.
   \item \textit{Sexual dimorphism} - where male and females have dissimilar and distinct features, such as \textit{Oryctes nasicornis} (European rhinoceros beetle).
    \item \textit{Variability in orientation and stance:}  resulting in different features being visible at varying view angles also makes identification challenging. 
    \item \textit{Multiple insects and pests in the image frame:} Multiple individuals of the same species in an image can make identification challenging as closely clustered insects can exhibit different features due to pose, orientation, and occlusion in comparison to features exhibited by an individual.
    
\end{enumerate}
The past decade has also seen efforts to harness the broader public to collect large datasets of scientific utility. Such citizen science efforts have recently produced large, diverse, high-quality, community-usable datasets~\cite{JFT,swag} that serve as the foundation for building automated insect classifiers. These data collection efforts, like iNaturalist~\cite{inat_dataset}, exhibit several desiderata: iNaturalist leverages the collective strength of its users (the crowd) for data curation. The user community is responsible for identifying the observations, with consensus from multiple users validating each identification. Every successful identification enhances the communal knowledge pool, contributing to a broader understanding of global biodiversity. The iNaturalist dataset provides wide-ranging coverage of species across vast geographic areas. It is thoroughly documented and undergoes regular updates for accuracy and comprehensiveness. iNaturalist \cite{inat_dataset} consists of over 70 million images, with over 13 million images belonging to class Insecta. Given this large data set, a few additional challenges have to be resolved to create a robust, automated insect classifier trained on such datasets:
\begin{enumerate}
\setcounter{enumi}{8}
\setlength{\parskip}{0pt}
\setlength{\itemsep}{0pt plus 1pt}
    \item \textit{Number of insect species imbalance among species categories}: i.e. large variations in the number of images in each insect species categories across different insect species can make training non-trivial. 
    \item \textit{Robust} identification, either by abstaining from classification when uncertain or by providing confidence bounds on predictions, can ensure enhanced trustworthiness when deployed in the wild.
\end{enumerate}

Here, we use recent advances in self-supervised training \cite{swav}, inter- and intra-domain transfer learning, out-of-distribution detection \cite{saadati,hendrycks2016baseline,lee2018simple,liu2020energy}, and conformal predictions \cite{angelopoulos} to train a robust insect classifier, called InsectNet. The classifier exhibits $>96\%$ classification accuracy on a large set (2526 insect species categories) of agriculturally and ecologically relevant insect and pest species. In contrast, the previous best classifier trained on the Insecta class 
 (using the 2017 iNaturalist test dataset~\cite{inat_dataset}) exhibited a top-1 accuracy of 77.1\%. InsectNet demonstrates success in overcoming each of the challenges outlined above.

\section{Results}
We focus on 2526 agriculturally and ecologically  important insect species (see SI: Section 1.1 for a list of insect species, number of images per species, and taxonomic information). 
We describe the technical workflow of our approach. Additional details are provided in SI: Section 2

\subsection{Technical workflow of InsectNet}
\textit{A. Citizen science collected dataset}: We selected a subset belonging to the class Insecta from the full iNaturalist dataset ($> 70M$ images). This subset consisted of $13M$ insect images belonging to around 100,000 distinct insect species. We further filtered this data to identify a subset of 2526 species categories of insects consisting of both beneficial insects and harmful pests that are agriculturally and ecologically relevant. The beneficial insects consist of pollinators,  parasitoids, and predator species. This dataset, comprised of $6M$ images, has been curated and quality checked by domain experts to ensure accurate species labels. The labeled images span 17 insect orders, with the order Lepidoptera containing the highest number of species (1430 species) and Zygentoma containing the lowest (3 species). Within these orders, some charismatic species, such as the \textit{ Danaus plexippus} (monarch butterfly) from the order Lepidoptera, have as many as $136,000$ images. In contrast, other insect species, like the  \textit{Nisitrus vittatus} (common bush cricket) from the order Orthoptera, have as few as 38 images. This is nearly three orders of magnitude variation in data availability and highlights a significant data imbalance challenge~\cite{he2009learning, cao2019learning} for training deep learning models (challenge \#9 above). 

\begin{figure*}[ht!]
\centering
  \includegraphics[ width=16cm]
  {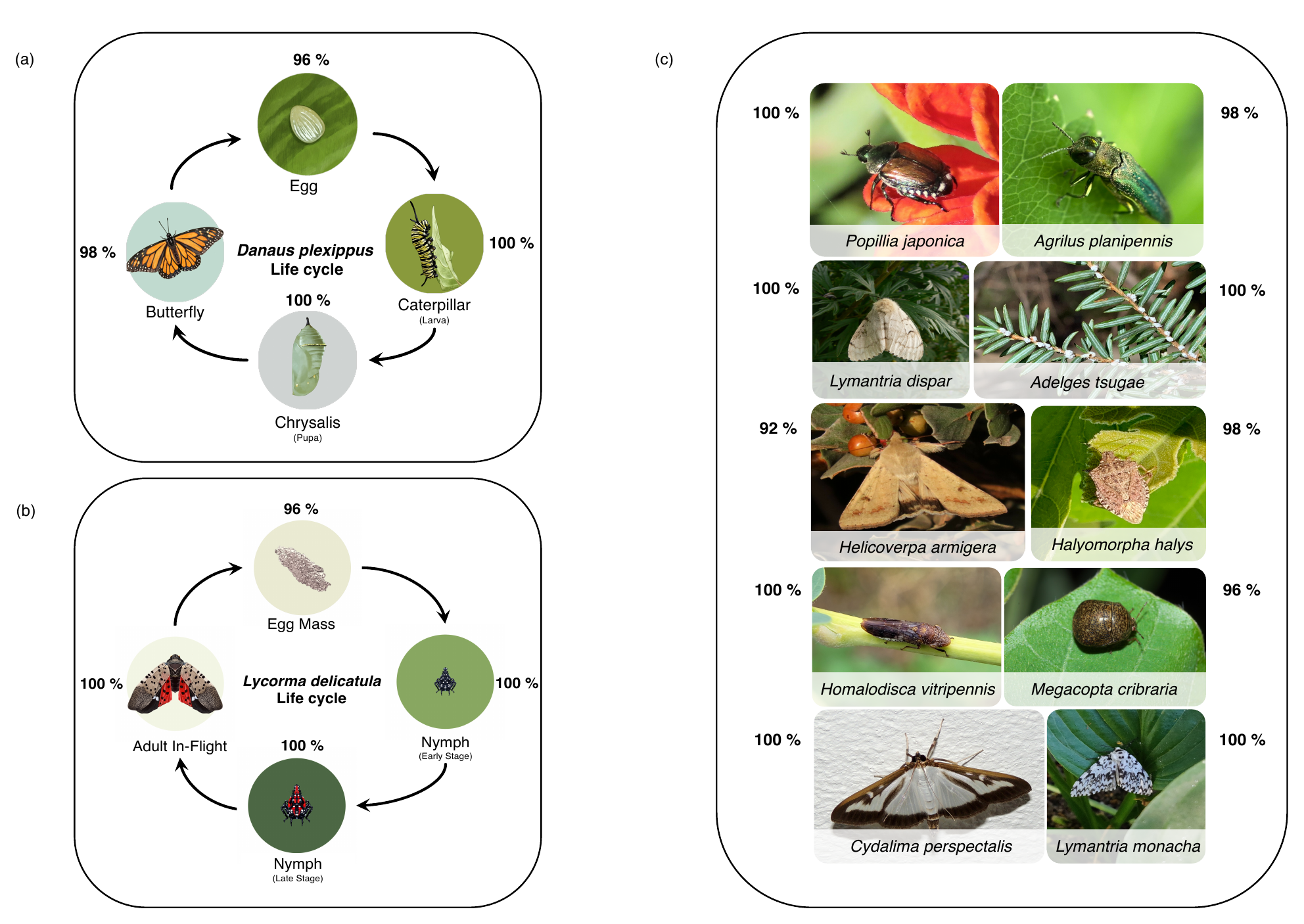}
  \caption{InsectNet is able to accurately identify insect species across the life cycle stages. Top left: charismatic species \textit{Danaus plexippus} (Monarch butterfly), Bottom left: an invasive species  \textit{Lycorma delicatula} (Spotted lanternfly). Right panel: Examples of the ability of InsectNet to accurately identify several invasive pest species. }
  \label{Fig:Invasive_LifeCycle}
\end{figure*}

This dataset comprises insects of varying sizes, from the smallest size species such as \textit{ Aphis nerii} (Oleander aphid or sweet pepper aphid) ranging from 2-3 mm to larger ones like the \textit{Hyalophora cecropia} (Cecropia moth), the largest moth in North America, with a wingspan reaching 15-20 centimeters. We utilize ten images per species from the iNaturalist 2021 dataset for testing and validation. Additionally, to ensure the statistical significance of reported per-insect species accuracies, we additionally collected and evaluated the performance of InsectNet on 50 public domain web images for all insect species depicted in Fig~\ref{Fig:Invasive_LifeCycle}, and Fig.~\ref{Fig:Intra-Inter Species}

\textit{B. Label-free self-supervised learning (SSL)}: Training accurate machine learning models require the availability of annotated datasets -- for instance, datasets where each insect image is tagged with a species name, or \textit{label}. Providing accurate labels for a large dataset is currently the most significant bottleneck in training accurate ML models, especially when label creation (or checking) requires expert knowledge. We utilize SSL approaches \cite{swag, swav}, which enable a model to initially learn useful features of a dataset without the need for any labels. Subsequent fine-tuning is then performed using a \textit{smaller labeled} dataset and has been shown to produce high-performing models~\cite{nagasubramanian2022plant}.

We perform an extensive series of training on several model architectures (RegNet, ResNet, see Supplementary: SI: Section 2.2) and report the impact of SSL pre-training across two performance axes: (a) The \textit{amount of unlabeled data} used for pre-training has a substantial impact on final classification performance. SI: Table S1, describes the impact of systematically increasing the amount of unlabeled data by 200X. These results quantitatively illustrate the value of citizen science collected data, with SSL approaches leveraging them even if such datasets are available without labels or when the labels are incomplete or noisy. (b) The \textit{number of pre-training campaigns} matters. That is, `daisy-chaining,' a model's pre-training on a sequence of different datasets or pretext tasks helps improve the final model performance. In SI: Table S2, we empirically show that classification models learn better latent representations when their model weights are sequentially trained across multiple datasets. Our best model consisted of one campaign of pre-training on a very large \textit{non-insect} dataset, followed by a second campaign of SSL pre-training on the \textit{insect} dataset, followed by final finetuning on \textit{labeled data} (See SI: Fig S1). This is a corollary to the first point (the amount of unlabeled data matters) by extending the approach to utilizing out-of-domain large datasets for pre-training (or rather pre-pre-training). We evaluated model performance along these performance axes to identify our best-trained insect classifier. This classifier exhibited a $96.4\%$ classification accuracy, with a $94\%$ mean per-species accuracy. The classification accuracy histogram for all the 2526 species categories exhibits a very small tail, suggesting that only a small fraction ($3.40\%$) of the species categories have a prediction accuracy of less than $80\%$ (See SI: Fig S2 for the histogram plot). Many of these species with lower accuracy possess less than 1000 images per species in the training dataset. We also note no correlation between insect size and prediction accuracy.   

\textit{C. Improving the prediction accuracy of species with a low number of images in the database (i.e., low sample size)}: We use an approach that transfers knowledge from high-accuracy categories with numerous examples to enhance the learning of low-accuracy categories with fewer examples. AlphaNet is a wrapper model that operates \textit{post hoc} on top of the insect classifier without requiring any retraining ~\cite{alphanet}. We demonstrate that AlphaNet significantly improves the prediction accuracy of low-accuracy species while retaining the overall prediction accuracy of the classifier. AlphaNet shifts the tail of the per-species accuracy histogram toward higher accuracy levels. In particular, the average accuracy of the low-accuracy species improved from $79.7\%$ to $87.6\%$, with only a $1.3\%$ drop in the overall classification accuracy (from $96.4\% \rightarrow 95.1\%)$. This ensures that almost all species in our insect species classifier exhibit a per-species prediction accuracy greater than $80\%$. This strategy addresses challenge $\#9$.

\textit{D. Improving trustworthiness of the model}: To ensure the robust performance of InsectNet in the wild, we wrap around two additional features to our classifier. First, we ensure that InsectNet avoids making predictions when confronted with low-resolution, blurred, or confusing images. This provides guardrails against potentially catastrophic consequences, for instance, the misclassification of an unseen insect species (say, belonging to an invasive species) as a benign insect species; or the misclassification of images belonging to a non-insect category (say, very small red color berry) as insects (say, lady beetles). We do this by wrapping around an out-of-distribution (OOD) detection algorithm around the classifier. The algorithm uses an energy-based metric (Refer SI: Section 4.1 ) to flag images that deviate significantly from the data distribution that the classifier is trained on (See SI:Fig S3 and S4 that depicts how the energy value can be used to distinguish in-distribution and OOD images). Our empirical analysis indicates that the $6M$ dataset exhibits a diverse set of imaging conditions making OOD detection a useful strategy --- yet another indicator of the power of citizen science data (See SI:Fig S5 that illustrates the results of InsectNet on OOD samples). Second, we use a conformal prediction approach to produce prediction sets, rather than a single species category, with rigorously guaranteed confidence (set to $\geq 97.5\%$). The prediction sets become larger when the classifier is increasingly uncertain of its prediction. Both these features provide a graceful way for human intervention and subsequent decision-making, thus resolving challenge~$\#
10 $. These features also allow quantitative feedback to direct citizen science data collection efforts for insect species where InsectNet underperforms.
    
\textit{E. Democratized access and streamlined MLOps}: The classifier is publicly available and hosted on a server : https://insectapp.las.iastate.edu. We also provide access to the trained model weights and quantized versions of the model that can fit into edge devices. Additionally, we provide access to all the MLOps workflows to enable the Ag community to adopt and leverage these approaches. In particular, to streamline the data wrangling process, we created a workflow tool, iNaturalist Scalable Download (iNatSD), that allows users to intuitively download customizable datasets of high-quality images of organisms in an ML-analysis-ready format. 

We next systematically evaluate the classifier against the challenges articulated in the introduction.

\subsection{InsectNet performance on challenges}
\textit{Challenge $\#1$, Large number of insect species}: The InsectNet model was extensively trained on a dataset of 13 million insect images encompassing numerous species, followed by fine-tuning on 6 million images belonging to 2526 insect pest species. Including a diverse array of insect species during training was imperative for ensuring InsectNet's high per-species category accuracy and robustness.


\textit{Challenge $\#2$, Metamorphosis (multiple life cycle stages)}: insects and pests species go through metamorphosis during their life cycle, which refers to the process of profound physical (color, shape, and structure) and developmental stage (egg, larva, pupa, nymph, and adult) transformation. Charismatic insect species like monarch butterflies exhibit complete metamorphosis and go through four distinct stages: egg, larva, pupa, and adult. We use this example to demonstrate that the classifier can successfully identify monarch butterflies at different life stages with high accuracy, see Fig~\ref{Fig:Invasive_LifeCycle}, and confidence scores. Another example is incomplete metamorphosis in SLF, which is an invasive species that transitions through three life cycle stages: egg, nymph, and adult. InsectNet can identify all three stages with high accuracy, see Fig.~\ref{Fig:Invasive_LifeCycle}a,b. The ability to identify insects early in their life cycle is especially important for efficient and early mitigation efforts and for preventing the establishment of SLF in new regions. In the case of SLF, early identification of egg masses on tree trunks, furniture, and buildings can help mitigation by chipping egg masses and destroying them \cite{egg_mass}.

Invasive pests species are of significant concern for agriculture as these non-native invasive species can cause great harm to horticulture and agriculture crop species, forest tree species, and urban green landscapes. The USDA National Invasive Species Information Center~\cite{terrestrial_invertebrates} lists invasive pest species that seriously threaten various food grain crops, vegetable, fruit, tree, and shrub species. Our model is able to accurately identify a large set, see Fig.~\ref{Fig:Invasive_LifeCycle}, including (\textit{Lycorma delicatula} (spotted lanternfly), 99\%); (\textit{Helicoverpa armigera} (Old World bollworm), 92\%); (\textit{Popillia japonica} (Japanese beetle), 100\%);  (\textit{Megacopta cribraria} (Kudzu bug), 98\%); (\textit{Halyomorpha halys} (Brown marmorated stink bug), 100\%); ~(\textit{Homalodisca vitripennis} (glassy-winged sharpshooter), 100\%); (\textit{Agrilus planipennis} (Emerald Ash borer ), 98\%); (\textit{Adelges tsugae} 
 (Hemlock woolly adelgid ), 96\%);  (\textit{Lymantria dispar} (Spongy moth), 100\%);  (\textit{Lymantria monacha} (Nun moth), 100\%); and (\textit{Cydalima perspectalis} (Box tree moth ), 100\%). Accurate identification of these invasive species at ports of entry and geographic borders can prevent the escape and spread of these invasive species into new geographic regions.

\begin{figure*}[h!]
\centering
  \includegraphics[width=\textwidth]{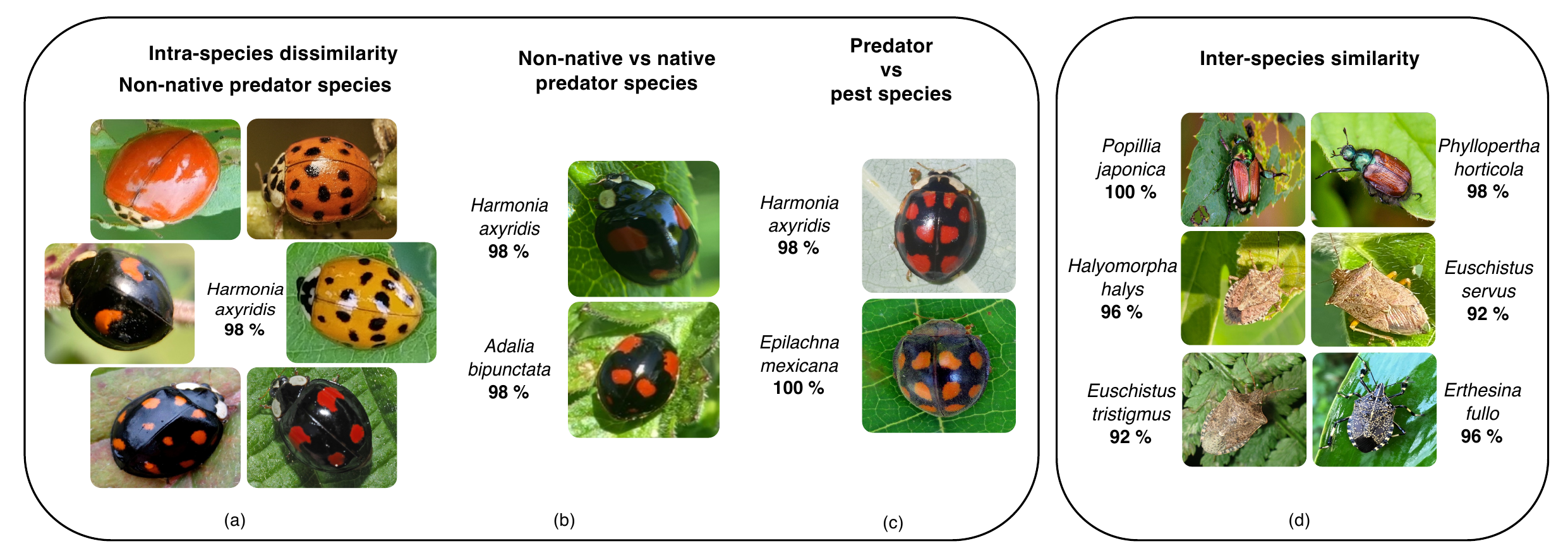}
  \caption{
  InsectNet can identify (a) intra-species dissimilarity of non-native  predator species \textit{Harmonia axyridis}(Asian lady beetle), (b) the difference between predator species of non-native Asian lady beetle and native beetle species \textit{Adalia bipunctata}(two-spotted lady beetle) (c) the difference between non-native predator species Asian lady beetle and pest species \textit{Epilachna mexicana}(Mexican bean beetle)  exhibiting similar features (all pattern variations not shown in the figure) (d) examples of inter-species similarity in case of look-alike beetles \textit{Popillia japonica} (Japanese beetles) and \textit{Phyllopertha horticola} (garden chafer) and different kinds of stink bug species.}  
  \label{Fig:Intra-Inter Species}
  \end{figure*}

\textit{Challenge \#3: Intra-species dissimilarity} in insect classification refers to the degree of dissimilarity among the members of an insect species. We showcase an example in this category  belonging to Coccinellidae family, \textit{Harmonia axyridis} (Asian lady beetle) which is a non-native species (Fig.~\ref{Fig:Intra-Inter Species}a). The variations in color and pattern exhibited by members make classification by non-experts nearly impossible; however, our classifier can successfully recognize six variations of the Asian lady beetle (accuracy $98\%$).

The Asian lady beetle was intentionally introduced into US regions with a lack of natural predators to regulate the population of soft-bodied pests like aphids, mealy bugs and scale insects etc.~\cite{asian_lady_beetle}. While both native and non-native lady beetle species are important as predators, the non-native species (Asian lady beetle) has become a nuisance as it out-competes native species such as \textit{Adalia bipunctata} (two spotted lady beetle), resulting in biodiversity loss \cite{biodiversity}.  Other detrimental consequences of the Asian lady beetle in North America includes causing harm to fruit crops and acting as a home intruder~\cite{koch}.

\textit{Challenge \#4: Inter-species similarity:} Different insect-pest species can look similar due to similarity in color and pattern. For instance, more than 500 species of lady beetle are reported in the U.S., making identification challenging; Our classifier performs well on this challenge, see Fig.~\ref{Fig:Intra-Inter Species}b,c, with accuracy ranging from $96\%$ to $100\%$. InsectNet can differentiate between predator species of non-native Asian lady beetle and native beetle species \textit{Adalia bipunctata} (two spotted lady beetle) along with the ability to differentiate between predator (\textit{H.axyridis}) vs. pest species of lady beetle (\textit{Epilachna mexicana} (Mexican bean beetle)). Accurately differentiating between visually similar species is important for timely mitigation, especially when harmful insect species look like beneficial predatory species. InsectNet can also accurately differentiate between two look alike beetles, see  Fig.~\ref{Fig:Intra-Inter Species}d. \textit{Popillia japonica} (Japanese beetle) and \textit{Phyllopertha horticola} (garden chafer) have very similar overall appearances, and experts differentiate them using subtle differences in physical features. Additional examples illustrated in Fig.~\ref{Fig:Intra-Inter Species}d include differentiating between \textit{Euschistus servus} (Brown stink bug; insect-pest, native to U.S.), \textit{Halyomorpha halys} (brown marmorated stink bug; insect-pest, invasive in U.S.), \textit{Euschistus tristigmus} (dusky stink bug; insect-pest, native to U.S.) and \textit{Erthesina fullo} (Yellow spotted stink bug; insect-pest, invasive in U.S.). Polyphagous invasive pest species like the Brown marmorated stink bug is a global pest that harms over 170 plant species ranging from vegetable, fruit, food grain, and flower crop species \cite{global_invasion}. However, the look-alike predator species of stink bug, \textit{Podisus maculiventris} (spined soldier bug), preys on insects like caterpillars, aphids, and beetles, thereby controlling pest populations in gardens and agriculture. The ability to differentiate between a pest and a beneficial species is critical for appropriate mitigation without unnecessarily harming local biodiversity.

\begin{figure*}[h!]
\centering
  \includegraphics[width=0.9\textwidth]{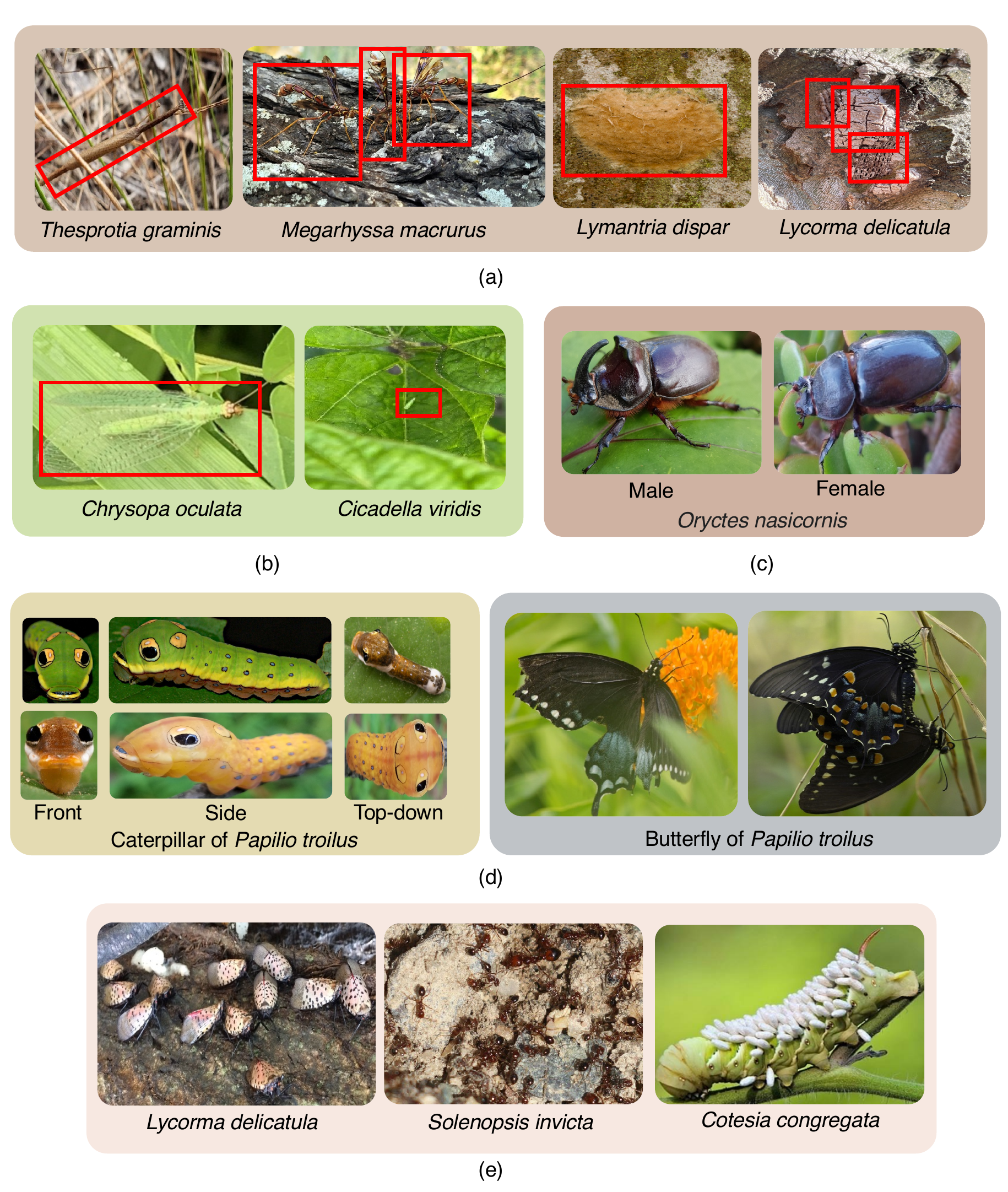}
  \caption{InsectNet can accurately classify under various challenging conditions: (a) camouflaged insects (brown insect on brown background), (b) camouflaged insects (green insect on green background),(c) sexual dimorphism (d) different poses and orientations, (e) multiple insects of same species in an image frame.}
  \label{Fig:Backgrounds}
\end{figure*}

\textit{Challenge \#5: Insect camouflage and diverse background}: Numerous insect species have patterns or colors that camouflage with the background, like a green insect on a green leaf or a brown insect on a piece of wood. Insects have evolved a variety of adaptation mechanisms that helps them blend in with their surroundings, resulting in a camouflaging effect to avoid predators and increase the chance of survival \cite{insect_physiology}. However, this camouflaging effect makes it challenging to identify the insect in their habitat \cite{camouflage_segmentation}. Our classifier performs well even for insect images in camouflaging backgrounds and small foreground-large backgrounds to produce reasonable predictions in such challenging cases, see Fig.~\ref{Fig:Backgrounds}a,b, with prediction accuracy ranging from 90 - 100\%. Examples illustrated include the \textit{Thesprotia graminis} (American grass mantis) which is a brown insect in a brown background, \textit{Megarhyssa macrurus} (long tail giant Ichneumonid wasp) that camouflages with tree bark, egg masses of the Spongy Moth, which bear a resemblance to a sponge, and the brown egg clusters of the Spotted Lanternfly; and example of green insect on green background, \textit{Chrysopa oculata} (green lacewing) and Cicadella viridis (green leafhopper) which is a very tiny green insect against a green leaf.

\textit{Challenge \#6: Sexual dimorphism}: In numerous insect species, male and females have dissimilar and distinct features. For example, the \textit{Oryctes nasicornis} (European rhinoceros beetle) is a species of beetle native to Europe, western Asia, and northern Africa, and has a large size, reaching up to 4 cm in length ~\cite{rhino_beetle}. The differences between male and female European rhinoceros beetles are not very pronounced, but there are some noticeable physical differences between them. The male has a characteristic horn on its head, similar to that of a rhinoceros (Fig.~\ref{Fig:Backgrounds}c). While it is not considered a major pest and primarily feeds on decaying matter, it still causes losses as adult feeds on the sap of a variety of trees, while the larvae feed on the roots of these trees and can cause significant damage to young trees. Our classifier is able to correctly identify images belonging to this species, irrespective of sex.

\textit{Challenge \#7: Variability in insect orientation and stance}: The example of the \textit{Papilio troilus} (spicebush swallowtail butterfly), see Fig.~\ref{Fig:Backgrounds}d, demonstrates the complexity of classification across the instar larvae and adult, where images are often taken from varying stance and pose (front, top, side). Our classifier correctly identifies the insect species corresponding to these images. It also correctly identifies the butterfly with broken wings, as well as an image of two butterflies with wings closed. 

\textit{Challenge \#8: Multiple insects and pests in the image frame}: In the wild, particularly for smaller-sized insects, multiple insects (at various life stages) are often present in the same image. Our classifier is able to make successful predictions across a variety of species, including \textit{Lycorma delicatula} (Spotted lanternfly), and \textit{Solenopsis invicta} (Red imported fire ant) with an accuracy of 100\% and 90\%, respectively, see Fig.~\ref{Fig:Backgrounds}c. A fascinating example of this ability is in the right image of Fig.~\ref{Fig:Backgrounds}c, which shows the \textit{Cotesia congregata} (parasitoid Braconid wasp ) cocoons on late-stage \textit{Manduca sexta} (tobacco hornworm) larva. The female braconid wasp lays her eggs inside the body of hornworm larva using a long, needle-like ovipositor. The eggs hatch into tiny larvae, which feed on hornworm body tissue, eventually killing it. Once the larvae have completed their development, they emerge from the host body and spin cocoons on the surface of the hornworm's skin. InsectNet successfully identifies multiple Braconid wasp cocoons on the hornworm body surface.

\begin{figure*}[h!]
\centering
  \includegraphics[trim={0 15cm 0 6.5cm},clip, width=\textwidth]{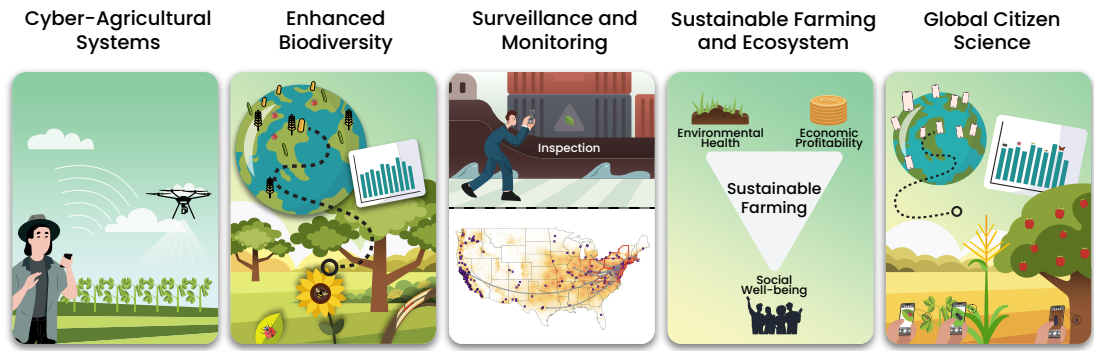}
  \caption{Impact of InsectNet and the democratized workflow. \textit{Agriculture}: Automated spatiotemporal resolved identification of pests can produce advances in cyber-agricultural systems and decision support. \textit{Biodiversity maintenance and enhancement}: InsectNet can enable rigorous and automated quantification of biodiversity gain/loss and help direct investments and policy. \textit{Trade}: InsectNet-like models can be deployed at the port of entries to automatically detect invasive species and monitor the spread of invasive species (taken from Ref~\cite{huron2022paninvasion}Fig 4a). \textit{Education}: InsectNet can be incorporated into extension training, as well as across the K-12 education ecosystem.}
  \label{Fig:FutureImpact}
\end{figure*}

\section*{Discussion}
We demonstrate the power and effectiveness of citizen science data (i.e., iNaturalist) to solve a significant challenge in crop production now and in the future. The well-curated dataset, coupled with a sequence of sophisticated deep learning tools, significantly improved our ability to automatically identify insect species in their various growth stages and enable deployment in the wild. Our classifier, InsectNet, produces consistently accurate predictions, even for challenging insect species that hobby gardeners, farmers, and plant scientists have difficulty identifying under field conditions. The classifier begins to fill the pressing need to identify insect-pest infestation at an earlier stage due to the increased population of invasive species and their improved ability to proliferate rapidly. We show that InsectNet can identify insects as early as when their eggs are laid, which would help with early mitigation.

InsectNet robustly identifies beneficial and harmful insect species including invasive insect-pest species and opens up a diverse set of unique opportunities, as illustrated in Fig.~\ref{Fig:FutureImpact}. This includes monitoring and surveillance of pests at international crossings/border inspections and tracking the domestic spread and movement of insect species. Historically, the identification of insects has been primarily focused on the adult, nymph, and larval stages. However, with advancements in image-based phenotyping, it is now possible to identify the eggs of pests and invasive species as well. This is particularly important in the case of invasive pests like \textit{L.dispar}, whose egg mass bears a sponge-like resemblance. The public's ability to recognize these egg masses is critical for pest control, as this insect spends around 10 months of its life cycle in the egg stage. InsectNet can be a tool for maintaining and enhancing biodiversity - for pollinators and other beneficial insects. InsectNet opens up several follow-up possibilities, including automated identification of insects in images and videos, and better integration of these technologies in Integrated Pest Management (IPM) and Climate Smart Pest Management (CSPM). Additionally, one could integrate such workflows in cyber-agricultural systems spanning (a) sensing - for example, through a smartphone (or edge computing) insect app; (b) modeling - for example, through digital twin-based prediction of establishment and movement; and (c) actuation - for example, through ground robots and drone-based automated control/mitigation, for gain in efficiency, sustainability, and profitability. We envision that such workflows will enable a transition from regional citizen science dataset collection to a global citizen science effort agnostic to country size, location, resources, and economy. In the context of insect species work, it can involve collective and coordinated activities on data collection for major and minor insect species, and use these resources for sustainable farming and ecosystem maintenance. Finally, we anticipate that this work (and the model weights) opens up efforts to (a) create fine-tuned models that are local to specific geographical regions and (b) extend to insect counting rather than just classification.  


\bibliography{sciadvbib}
\bibliographystyle{ScienceAdvances}

\noindent 

\noindent \textbf{Funding:} This work was supported by the AI Institute for Resilient Agriculture (USDA-NIFA \#2021-67021-35329), COALESCE: COntext Aware LEarning for Sustainable CybEr-Agricultural Systems (NSF CPS Frontier \#1954556), and Smart Integrated Farm Network for Rural Agricultural Communities (SIRAC) (NSF S\&CC \#1952045). Support was also provided by the Plant Sciences Institute. \\
\noindent \textbf{Author Contributions} 
AS and BG designed the project; SC, ZKD, NM, AS acquired the data; SC, MS, ZKD, JK, TJZ created new software and/or performed analysis; SC, AKS, SS, AS, BG interpreted data; SC, TZJ,   DM, AKS, AS, BG created first draft; all authors edited, reviewed, and approved the final draft.\\ 
\noindent \textbf{Competing Interests} The authors declare that they have no competing interests.\\
\noindent \textbf{Data and materials availability:} All data and model weights are publicly available. All links for data download are available in Supplementary Material. \\

\newpage

\section*{Supporting Information}

\setcounter{section}{0}
\section{Data}
\label{SI:Data}
\subsection{List of Insects}
\label{SI:InsectList}

InsectNet, our deep learning insect classifier hosted on the web app identifies a broad spectrum of insect species. The list of insect species that the app successfully identifies can be found :https://github.com/ShivaniChiranjeevi/Insect-Classifier/blob/main/classes.csv

\subsection{Dataset details}
{iNaturalist is a citizen science platform where users can upload labeled photographs of specific organisms. The iNaturalist Open Data project is a curated subset of the overall iNaturalist dataset that specifically contains images that apply to the Creative Commons license. It is a partnership between iNaturalist and Amazon, specifically created to aid academic research. The iNaturalist dataset is taxonomically relevant data for insect classification and identification problems. The Naturalist insect dataset is categorized into several hierarchical levels, including kingdom, phylum, class, order, family, genus, and species.  Insects belong to Kingdom Animalia and phylum Arthropoda. The phylum Arthropoda can be further classified into several subphyla, including subphylum Chelicerata (eg. spiders and mites), subphylum Myriapoda (e.g., centipedes and millipedes), sub-phylum Hexapoda (e.g. insects) and subphylum Crustacea (e.g., crabs and shrimp, etc.) The sub-phylum Hexapoda can be further divided into class Insecta which has 32 orders, and each insect order will have families, which is further classified into genera and at the species level depending on their distinctive attributes and traits ~\cite{orders}. The current classifier classifies insects at the species level which requires images at the last level in taxonomic classification.

\subsection{iNaturalist Scalable Download}
We created a workflow tool, iNaturalist Scalable Download (iNatSD), to easily download species images from the iNaturalist Open Dataset associated with a specific taxonomy rank. The tool utilizes the python Snakemake workflow manager to allow users to intuitively download customizable datasets of high-quality labeled images of organisms, and the ability to parallelize downloads based on the computational power of the machine. We used the tool to download all images of species under the rank class Insecta from the iNaturalist Open Dataset for use in our model. The complete dataset comprises a total of roughly 13 million images across 95 thousand different insect species at the time of writing. The images have a maximum resolution of 1024x1024, are in .jpg/.jpeg format, and total 5.7 terabytes. Among the 95 thousand insect species, we have used 2526 species that have been reported to be the most agriculturally and ecologically important species. This subset of insect classes contributes to ~6 million images in total. We choose to only use images identified as “research” quality grade under the iNaturalist framework, which indicates that the labeling inspection for the image is more rigorous than standard images, and has multiple agreeing identifications at the species level. 

\section{Multi-step training}
\label{SI:multi-step-training}

\subsection{Data Preparation}
Our model is trained under different settings using datasets of varying sizes: 66k, 660k, 2 million, and 6 million images. The 66k and 660k sets are balanced by class, while the 2 million and 6 million subsets are imbalanced. To validate and test our model, we used a total of 25260 images (10 images per class). These images were pre-processed by reshaping them into a 224 (height) x 224 (width) x 3 (number of channels) format and normalizing them. We utilized augmentation techniques to artificially increase the dataset size during training, which improves the model's generalizability and robustness. These techniques include geometric and color space transformations such as flipping, cropping, and adjustments to brightness or contrast. Our implementation utilized standard augmentation techniques such as horizontal flip and random erase, and we also adopted recent techniques such as CutMix and Mixup-Alpha, which have been demonstrated to enhance classifier performance \cite{cutmix,mixup}.

\subsection{Classifier Architecture Choices}
\label{architectures}
In our study, we utilized SSL pretraining to facilitate downstream classification and identify 2526 insect classes. Though the dataset sizes employed for pretraining differed in various experiments, we performed the end-to-end classifier finetuning on the balanced 660k data subset. Two different CNN architectures are used (ResNet and RegNet) and are explained in detail below.
\begin{enumerate}
\item ResNet: ResNet mainly addresses the vanishing gradient problem, which means with the network depth increasing, accuracy gets saturated  and then degrades rapidly \cite{resnet}. As we make the CNN deeper, the derivative when back-propagating to the initial layers becomes almost insignificant in value. To overcome this problem, ResNet uses skip connections from the previous layers. Among all the variations of ResNet models that differ in the depth of the network, we choose ResNet-50. It is a 50-layer convolutional neural network that can be utilized as a state-of-the-art image classification model. This model has been largely studied and explored for various dataset types. However, it is different from traditional neural networks in the sense that it takes residuals from each layer and uses them in the subsequent connected layers 
This model contains approximately 23 million trainable parameters.
\item RegNet: RegNet is an optimized design space developed by Radosavovic et al \cite{regnet}  where they explore various parameters of a network structure like width, depth, groups, etc. RegNet is derived after simplification from AnyNet, an initial space of unconstrained models which uses models like ResNet as its base. It is a type of deep neural network architecture used for image classification tasks. It is designed to be scalable, meaning that its architecture can be easily adapted to accommodate larger or smaller models, depending on the size of the dataset and computational resources available. In RegNet models, the network architecture is defined using a mathematical formula that specifies the number of filters (the neurons in the network) as a function of the resolution of the input image. This allows the network to be easily scaled up or down, without manually specifying the number of filters for each layer. RegNet models have been shown to achieve state-of-the-art results on several benchmark datasets, and are widely used in computer vision applications. They are attractive due to their scalability and efficiency and ability to learn high-level representations of images that are useful for classification tasks. By conducting many experiments where different parameter values are tested for the design space, they arrived at the optimized RegNetX or RegNetY models. It is an improved version of RegNet that has been optimized for both efficiency and accuracy. The "Y" in RegNetY refers to the network structure, which is shaped like the letter "Y." In this architecture, the network branches out from a central stem, with each branch processing a different level of information from the input image. This allows the network to learn multiple scales of features from the image, which can be combined to make more accurate predictions. In this paper, we use the RegNetY32 model belonging to the family of RegNetY models for our experiments which roughly has 145 million trainable parameters.
\end{enumerate}
The classifier is built using the PyTorch library and is available online at: https://github.com/pytorch/vision/tree/main/references/classification.

\subsection{Self-Supervised Pretraining}
\label{SSL_swav}
Self-supervised learning (SSL) is a type of machine learning technique that enables a model to learn from data without any human-annotated labels. In SSL, the model is trained to make predictions about the data by creating a task that is not directly related to the final objective. For example, the model may be trained to predict missing parts of an image or to cluster similar images. These algorithms can learn from vast amounts of unlabeled data, making them useful when labeled data is scarce or expensive to obtain. SSL learns to extract useful features from the data, which can be used for a variety of downstream tasks, improving the model's generalization ability. Additionally, SSL can significantly reduce the amount of labeled data, time and cost needed for training a classifier. Our proposal is to use SSL methods that only rely on unlabeled data to learn representations that could differentiate between different classes. Even with a large corpus of labeled data, SSL greatly benefits the accuracy of the downstream classification task by learning robust contrastive features, which could be used for classifying other datasets belonging to similar domains. An efficient pre-trained SSL model could prove to be very useful in various applications.
\newline
\newline
One effective SSL method is the Swapping Assignments between Views (SwAV), an online clustering-based SSL method \cite{swav}. SwAV works by creating different augmented views of an image at different scales and training the model to cluster together similar versions of the same image. The network learns representations by using a simple cross-entropy loss to enforce it to learn the cluster assignment code of one augmented view from the other. The idea behind this is that the augmented views originate from the same image and should contain the same features or class information. To ensure that all clusters contain approximately the same number of samples (equipartition constraint), the Sinkhorn-Knopp algorithm is used. SwAV has improved memory efficiency and top-1 accuracy compared to other contrastive SSL methods and was therefore adopted for our experiments.
\newline
\newline
Our experiments investigates the effect of scaling up model size, data size, and depth of pretraining on the downstream accuracy of a very complex dataset of fine-grained and highly imbalanced class distribution consisting of 2526 insect-pest classes. Our work builds upon the findings of earlier studies that explore the impact of pretraining on downstream tasks \cite{google_exploring}. We also draw from another study on the effects of varying upstream task's pretraining dataset size, domain of the data, and model size on downstream task accuracy\cite{mehdi}.  We aim to determine the optimal axes of pretraining and subsequent fine-tuning of the insect-pest classifier, leveraging existing powerful pretraining techniques and CNN architectures, to significantly enhance the performance of the classifier. The proposed recipe for optimal pretraining and fine-tuning shows promising results for the classification of the insect-pests with high mean-per-class accuracy.

\subsubsection{Impact of pretaining dataset size}
The original dataset pool contains 13 million images. Multiple subsets of sizes 66k, 660k, 2 million, and 6 million were extracted to test the effect of different data sizes during pretraining on the downstream classification. The CNN models are pretrained using SwAV along these different unlabelled dataset sizes and finally fine-tuned on a small 660k labeled dataset. The SSL pretraining helps the model learn in-domain feature representations without labels. These feature representations are used for the classification which utilizes a small labeled subset. We utilized varying subset sizes of dataset are used for pretraining. For the two different models used in our experiments, we observed that with increased dataset size, the mean-per-class accuracy and top-1 accuracy of the classifier increased as depicted in Table \ref{table:experiment_1}.
\begin{table*}[h!]
\centering
\caption{Performance of classifier with varying pretraining dataset sizes}
\subtable[ResNet Architecture]
{\begin{tabular}{ccc}
\textbf{Dataset Size} & \textbf{Top-1 Accuracy} & \textbf{Top-5 Accuracy} \\
\midrule
66k & 86.60 & 95.81 \\ \hline 
660k & 88.90 & 95.48 \\ \hline
2 Million & 88.26 & 96.93\\ \hline
6 Million & 89.25 & 97.18\\ 
\bottomrule
\end{tabular}}
\subtable[ RegNet Architecture]
{\begin{tabular}{ccc}
\textbf{Top-1 Accuracy} & \textbf{Top-5 Accuracy} \\
\midrule
89.20 & 96.97 \\ \hline 
89.39 & 97.06 \\ \hline
90.72 & 97.52\\ \hline
92.56 & 98.27\\ 
\bottomrule
\end{tabular}}
\label{table:experiment_1}
\end{table*}

\subsubsection{Impact of depth of pretraining}
The next axis of pretraining studied that affects the classification is the depth of pretraining. Depth of pretraining is defined as the number of times a  deep learning model is pretrained using datasets of different domains or strategies. There are two depths of pretraining
\begin{enumerate}
    \item Depth-1 Pretraining: Includes one level of SSL pretraining on in-domain insect-pest data
    \item Depth-2 Pretraining: Includes two levels of pretraining, a large scale, out-of-domain pretraining followed by SSL pretraining on in-domain data as depicted in Fig \ref{Fig:pipeline}.
\end{enumerate}
\begin{figure*}[ht!]
    
  \centering
  \includegraphics[width=18cm,height=4.5cm]{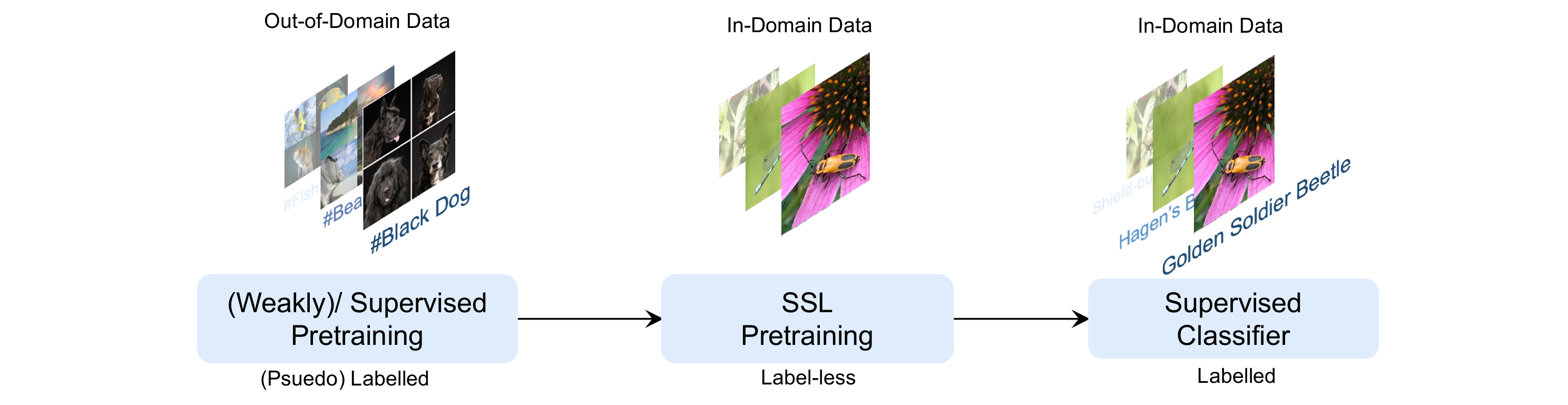}
  \caption{Flow diagram depicting the depth-2 pretraining and finetuning pipeline of our insect classifier }
  \label{Fig:pipeline}
\end{figure*}
We compare and contrast the different pretraining depths for two different model sizes while keeping the pretraining dataset size fixed for both models though the pretraining setting varies. While the final level of pretraining at both the depths includes SSL pretraining, the level-1 pretraining method for (2) is chosen for ResNet and RegNet is different for both the models and was chosen after experimenting with the various available pretraining strategies.
For ResNet, the level-1 pretraining for (2) uses out-of-domain ImageNet-21k \cite{imagenet} supervised pretrained weights whereas RegNet follows a weakly-supervised out-of-domain pre-training recipe called SWAG (Supervised Weakly from hashtAGs) which uses 3.6 billion Instagram images and their corresponding hashtags (Hashtag Supervision). It is weakly supervised as each image does not have a single well-defined label but a number of hashtags that may vary for each image and same images could be described with a different set of hashtags. The pretraining strategies chosen for both models provided the best accuracy among other supervised and self-supervised pretraining strategies. We compared the effect of different pretraining depths on the downstream classification. In addition to one level of SSL pretraining (Depth-1), using powerful large-scale pretrained models available to public prior to this (Depth-2) significantly boosts the classifier's performance, which is evident from Table \ref{table:experiment_2}. Depth-2 pretraining combines the benefits of large-scale out-of-domain dataset to learn generic coarse features and domain-specific fine-grained features.

\begin{table*}[h!]
\centering
\caption{Performance of classifier with varying pretraining depth}
\subtable[ResNet Architecture]
{\begin{tabular}{ccc}
\textbf{Dataset Size} & \textbf{Top-1 Accuracy} & \textbf{Top-5 Accuracy} \\
\midrule
1 & 83.18 & 93.86 \\ \hline 
2 & 85.34 & 95.27 \\ 
\bottomrule
\end{tabular}}
\subtable[ RegNet Architecture]
{\begin{tabular}{ccc}
\textbf{Top-1 Accuracy} & \textbf{Top-5 Accuracy} \\
\midrule
88.11 & 96.07 \\ \hline 
89.39 & 97.06 \\ 
\bottomrule
\end{tabular}}
\label{table:experiment_2}
\end{table*}

\subsubsection{Impact of size of model}
The two popular architectures ResNet-50 and RegNetY32 were chosen for comparison of varying number of trainable parameters and model sizes. ResNet-50 has roughly 23 million network parameters, whereas RegNet has 145 million network parameters. These two architectures are explained in detail in \nameref{architectures}.
We experimented with models having a different number of trainable parameters, ResNet-50 and RegNetY-32 with 23 and 145 million parameters each respectively. RegNet with increased model size gave notably higher accuracies than ResNet. The effect of scaling up the model size was more pronounced than the data size or depth of pretraining. The results are evident from Table \ref{table:experiment_1} and Table \ref{table:experiment_2}. 
\subsection{Details of best model}
Based on the experiments conducted where we trained multiple models with different SSL settings, we identified that the model trained with the highest pretraining dataset size, greater depth of pretraining and a model with more number of parameters provided us the best MPC. The model (RegNet) is pretrained at a depth of 2 on the highest subset of data available, 6 million insect-pest images from iNaturalist. This classifier provides an overall accuracy of \% and an MPC of \%. Fig \ref{Fig:app_histogram} shows the histogram of per-class accuracies of this model and it can be noticed that the majority of he insect classes have an accuracy between 90\% and 100\% whereas only a very insignificant percentage of classes have an accuracy lesser than 80\%. 
\begin{figure*}[ht!]
  \centering
  \includegraphics[width=10cm,height=8cm]{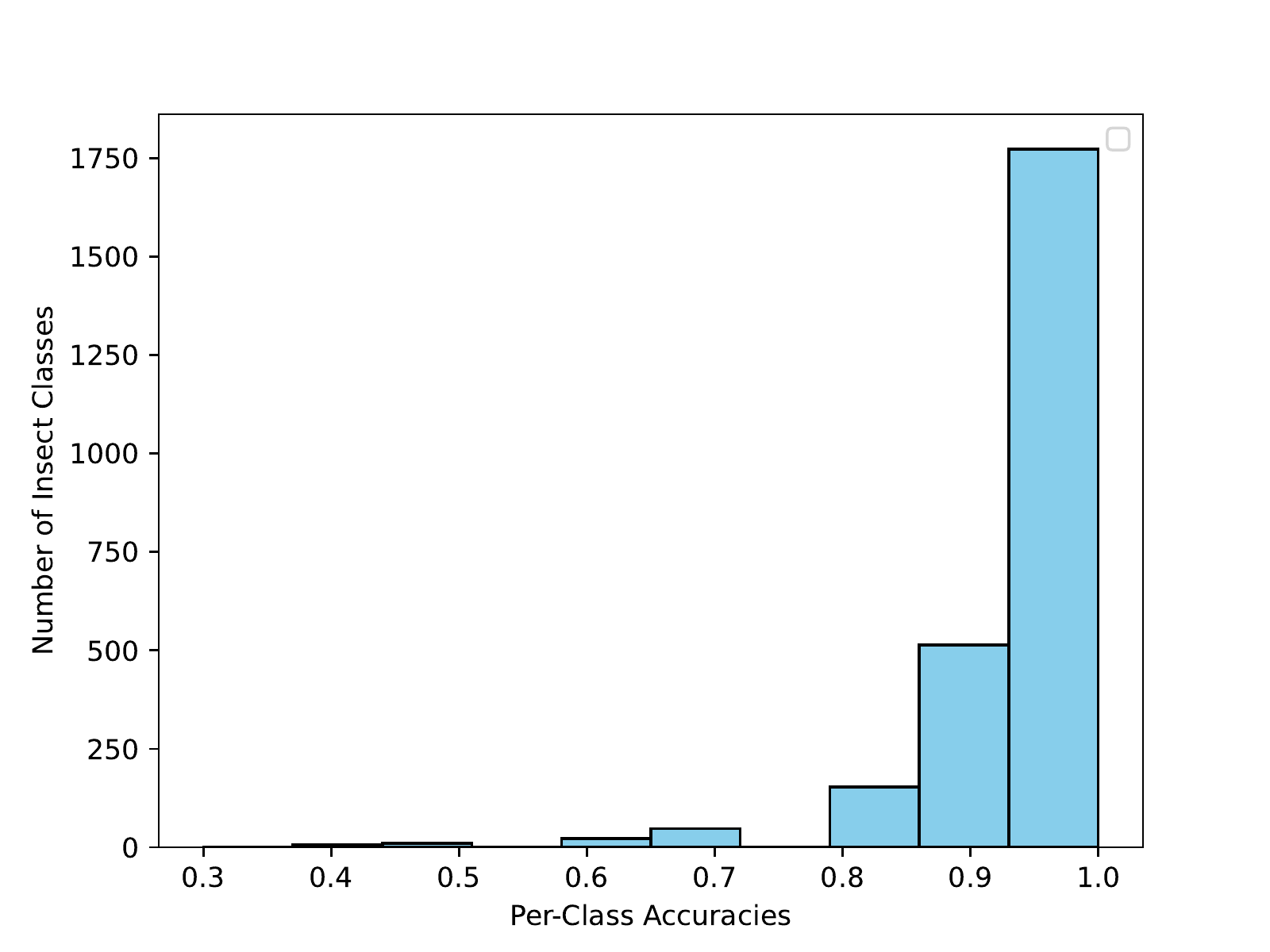}
  \caption{Histogram plot of per-class accuracies of 2526 insect classes. As evident from the plot, close to 70\% of insect classes have an 100\% accuracy.}
  \label{Fig:app_histogram}
\end{figure*}
\subsection{Compute effort}
All the experiments were carried out on the High-Performance Computing cluster, NOVA at Iowa State University on A100 GPUs. The resources and time required by SSL pretraining is significantly more than the downstream classifier finetuning and differ for the axes of pretraining like dataset size and model size. Spending a significant amount of compute resources for pretraining on large subsets of data like 6 million would yield great benefits as these pretrained models could be used for multiple downstream tasks in zero-shot or few-shot settings with almost no extra-added cost. 
\section{Fine tuning to account for data imbalance}
\label{SI:AlphaNet}

\subsection{AlphaNet}
AlphaNet is a framework that aims to address the challenge of long-tailed recognition, which is characterized by an unbalanced distribution of data with a long-tailed shape. In such scenarios, there is a lack of sufficient data for some classes, especially those in the tail of the distribution, resulting in poor performance for these classes during training and testing.
Few classes in our classifier suffer from low mean-per-class (MPC) accuracy. This could be attributed to poor diversity of images, low number of samples or poor quality of images in that particular class. It could be useful in improving the accuracy of classifying such rare and uncommon insect species that may belong to the tail of the distribution, where there is a lack of sufficient training data. By combining classifiers and adaptively adjusting them, AlphaNet could improve the accuracy of classifying these tail classes, which could be crucial in applications such as pest management. AlphaNet works post hoc on top of the existing insect classifier model without having to train it again \cite{alphanet}. It works to improve the low MPC accuracy of a few agriculturally important classes in our dataset without a significant drop in the overall accuracy of the baseline classifier. It transfers knowledge from models trained on higher accuracy classes to models trained on lower accuracy classes. It borrows information from high-accuracy classes similar in model representation to the poor-accuracy class and creates a new composition for that particular class. The classes are split into three categories. 'Few' split contains classes with accuracy lesser than 80\%, 'medium' split contains classes with accuracy between 80\% and 90\% whereas 'many' contains classes with accuracy greater than 90\%. The goal of AlphaNet is to considerably improve the accuracy of the 'few' classes while retaining the overall original accuracy of the classifier. 
\subsection{Impact of AlphaNet}
Our best classifier model pretrained on 6 million unlabelled insect data followed by finetuning with 6 million labelled dataset is fed into the AlphaNet for improving the accuracy on classes it recorded a low accuracy of.
It increased the overall accuracy of the 'few' or low-accuracy classes from 79.7\% to 87.6\% and there was only a 1.3\% drop in the overall classification accuracy.
\section{Improving trustworthiness}
\label{SI:Trust}
The trustworthiness of our insect classifier is significantly improved through the use of out-of-distribution detection and conformal predictions. Out-of-distribution detection refers to the process of identifying whether a given input belongs to a distribution that is different from the one the model was trained on. In other words, it helps to identify if the input is an outlier or an anomaly. This is important because if the input is outside the distribution, the model may not have the necessary information to make accurate predictions.

Traditional deep learning models output a single class label for each input image. They could often report a wrong prediction with high confidence which is not desirable and makes the model unreliable. Any classification prediction is associated with an uncertainty score. Conformal prediction uses these uncertainty scores to produce prediction sets that are guaranteed to have the ground truth label within them with a pre-determined or computed threshold value.  It can improve the reliability of predictions by providing uncertainty estimates that help decision-makers make more informed decisions.

By using both out-of-distribution detection and conformal predictions, the insect classifier app can be more trustworthy as it can detect when the input is outside the distribution of the trained model and provide a measure of confidence in the predictions. This is particularly important in agriculture where misidentification of insect pests can have significant economic and environmental consequences. By using these techniques, the app can help farmers make informed decisions about pest control measures and reduce the likelihood of misidentification.

\subsection{Out-of-Distribution Detection}
\label{OOD_section}

\subsubsection{OOD}
In machine learning, the out-of-distribution (OOD) problem occurs when a classifier is presented with samples that differ significantly from the training data distribution. For example, if a classifier is trained to classify cats and dogs, and is then presented with an image of a horse, it may fail to classify it correctly. In such cases, the classifier may produce a confident prediction that is wrong, which can lead to errors and trust issues for users of the classifier. Detecting this type of input is referred to as out-of-distribution (OOD) detection. Saadati et al.~\cite{saadati} conducted a comprehensive experiment on the combination of most well-known (Maximum SoftMax Probability, Mahalanobis algorithm) and state-of-the-art (Energy-based-algorithm) OOD algorithms and their performance on different insect classifiers and OOD dataset and proposed the best OOD detector for insect classification tasks  \cite{hendrycks2016baseline} \cite{lee2018simple} \cite{liu2020energy}.

\subsubsection{Energy-based OOD}
We applied the best OOD detector obtained from the experiments conducted by Saadati et al., an energy-based model (EBM), to our insect classifier \cite{saadati}. The energy-based algorithm is a mapping function from the logit value of the classifier to an energy level. The OOD detector then use energy level as an uncertainty score to determine if an input belongs to ID or OOD. The comparison of ID and OOD energy distribution is shown in Fig \ref{Fig:OOD_categories}. Based on the energy distribution of the ID and OOD data, an optimal energy threshold value is selected that distinguishes ID and OOD samples effectively. To ensure the robustness of our parameter estimation, we divide the data into 5 folds and get the average over 5 rounds. One advantage of energy-based OOD detection is that it can be applied to any classifier that produces a probability distribution over classes. This includes deep neural networks, support vector machines, and logistic regression, among others. Moreover, energy-based OOD detection can be used in conjunction with other techniques, such as conformal prediction, to further improve the reliability and trustworthiness of the classifier. Using a highly efficient energy-based OOD detection can help ensure that the classifier only makes confident predictions on insect samples that are similar to the training data distribution. This can improve the accuracy and trustworthiness of the app, and prevent false positives or false negatives that may harm crops or mislead farmers. Overall, the use of OOD detection techniques such as energy-based OOD can help improve the reliability and trustworthiness of machine learning classifiers in a wide range of applications.
\begin{figure*}[ht!]
  \centering
  \includegraphics[width=17cm,height=10cm]{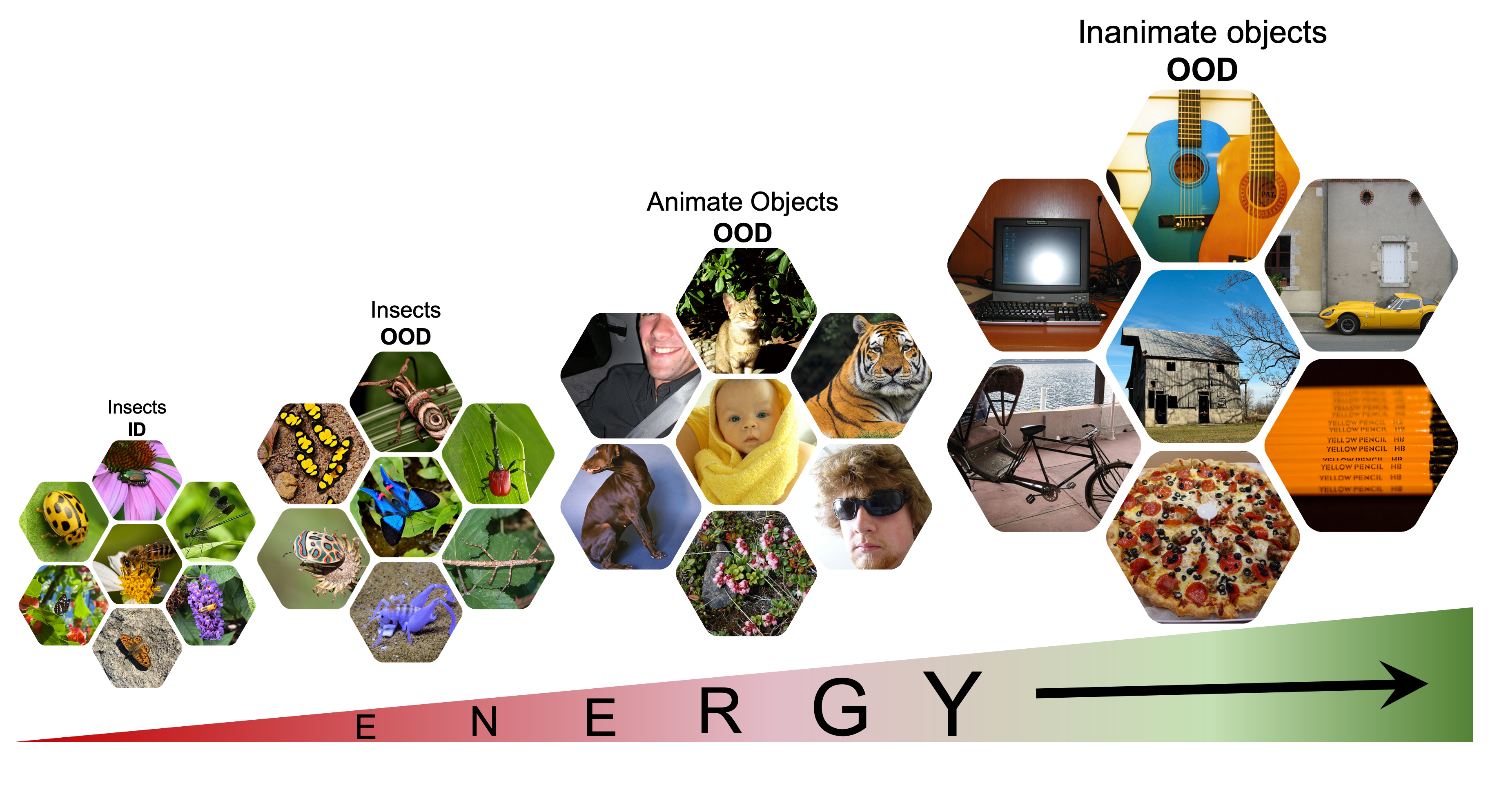}
  \caption{The energy value calculated by EBM outputs lower value for ID samples and higher values for OOD samples like insects that were not a part of the training data, non-insect living objects, and inanimate objects. The images used in the figure belong to ILSVRC dataset \cite{imagenet}.}
  \label{Fig:OOD_categories}
\end{figure*}
\subsubsection{OOD results}
For the energy-based OOD detection, we obtained an energy level of 11.49 as the best threshold to distinguish in-distribution to OOD data with an accuracy of 89.99\%. The graph in Fig \ref{Fig:violin_plot} depicts the energy distribution of ID and OOD data and a horizontal line is marked at the threshold energy level of 11.49 to visually inspect how well the classifier was able to distinguish between ID and OOD samples.. It was observed that OOD samples have higher energy values in comparison to the ID samples.
\begin{figure*}[ht!]
  \centering
  \includegraphics[width=6cm,height=5cm]{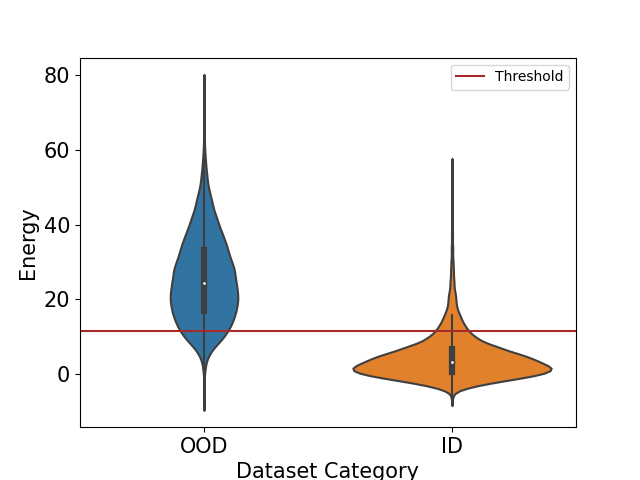}
  \caption{Energy distribution of ID vs OOD samples. The red horizontal line depicts the threshold energy value that differentiates the ID and OOD samples based on their calculated energy values.}
  \label{Fig:violin_plot}
\end{figure*}
\subsubsection{Implementation}
The InsectNet web app, which hosts our insect-pest classifier, has been equipped with an OOD detection feature. The OOD module does not only detect samples that have a significant deviation from the distribution of the training data but also other insect-pests that were not a part of the insect classes used during training. This is depicted in Fig \ref{Fig:OOD_examples}  (a). As shown in Fig \ref{Fig:OOD_examples} (b) and (c), if an image of an OOD sample, such as a baby's face or a golden fish , is uploaded, the model will display a prompt indicating uncertainty in its prediction and suggesting that it could be an OOD sample. 
\begin{figure*}[ht!]
  \centering
  \includegraphics[width=18cm,height=7cm]{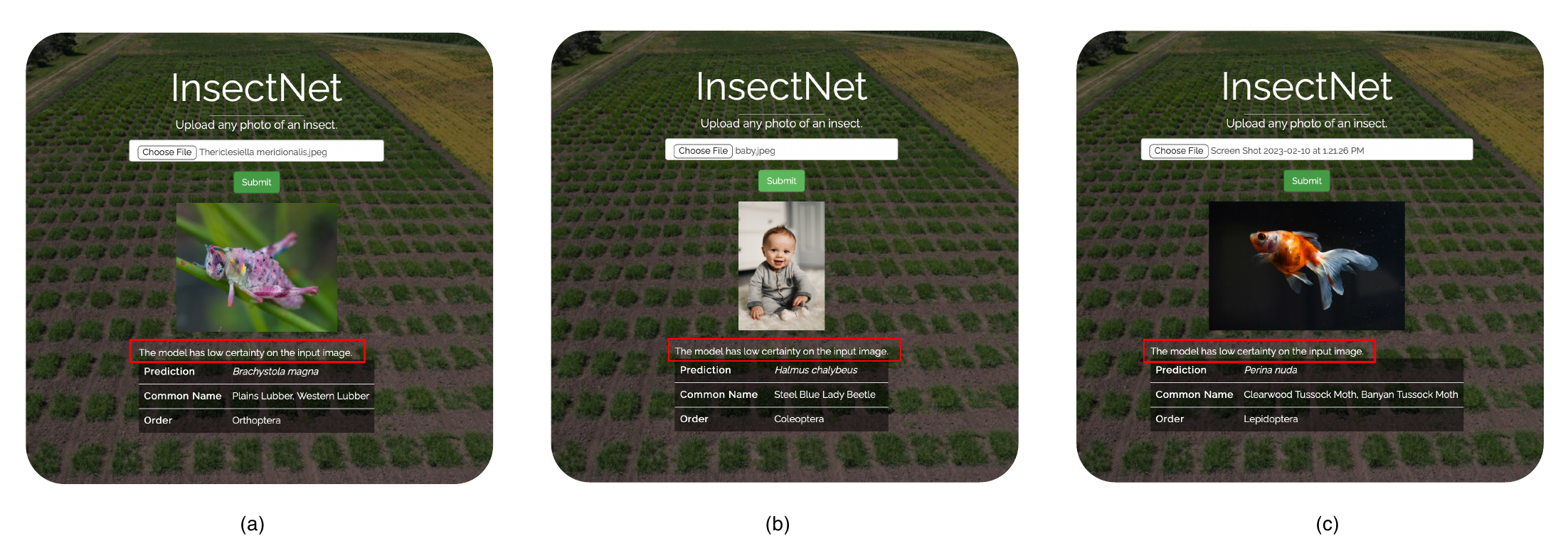}
  \caption{Images of OOD samples: (a) an insect class that does not belong to agriculturally important 2526 insect classes, (b) a human baby, (c) a golden fish  }
  \label{Fig:OOD_examples}
\end{figure*}
\subsection{Conformal Predictions}
Conformal prediction is an extrusive algorithm used as a wrapper around machine learning algorithms to provide a measure of uncertainty for each prediction. We used the conformal prediction framework implemented by Angelopoulos et al. \cite{angelopoulos}. For our analysis, we divided the Insecta validation set into two equal sizes of 12630 for training the conformal prediction algorithm and testing it. Let the classifier model's logit function be denoted by \textit{\^{f}}. We then implemented the following steps to obtain a conformal prediction set for our InsectNet predictions}:
\begin{enumerate}
    \item  We first chose a heuristic notion of uncertainty which in our case is the softmax value (denoted by $p$) of the function \textit{\^{f}}.
    \item We considered a threshold $\alpha$ as a probability that the true label belongs to our conformal prediction set. In our case, $\alpha = 0.025$.
    \item For each sample $i$ in our calibration set, we computed the uncertainty score of softmax of the true label, $s_i$ = 1 -$p_i$. 
    \item We sorted the conformal scores in increasing order. 
    \item We calculate  \textit{\^{q}} = $\lceil (n+1)(1-\alpha) \rceil$ / n and chose $\hat{q}$-quantile of the scores $s_i$, where n is the number of data points in the calibration set. 
    \item For new test samples, we include all classes in our conformal prediction set whose conformal score is lesser than $\hat{q}$-quantile.  
\end{enumerate}

Conformal predictors provide two types of guarantees. First, they are valid because they control the probability of making incorrect predictions within a given confidence level. Second, they are efficient because they do not require any assumptions about the underlying distribution of the data.
\subsubsection{Results}
As we discussed in the previous subsection, there are situations in which the model is uncertain between several predictions during the test. In these scenarios, it will be beneficial if the model returns all possible labels with their level of certainty. Conformal prediction is an approach for creating a set of potential labels such that the probability of the actual label belonging to the set is greater than a threshold. The size of the prediction set can vary and depends on the degree of similarity between the new data point and the training data. For our analysis, we used a threshold of 97.5\%.
\subsubsection{Implementation}
\begin{figure*}[ht!]
  \centering
  \includegraphics[width=10cm,height=5cm]{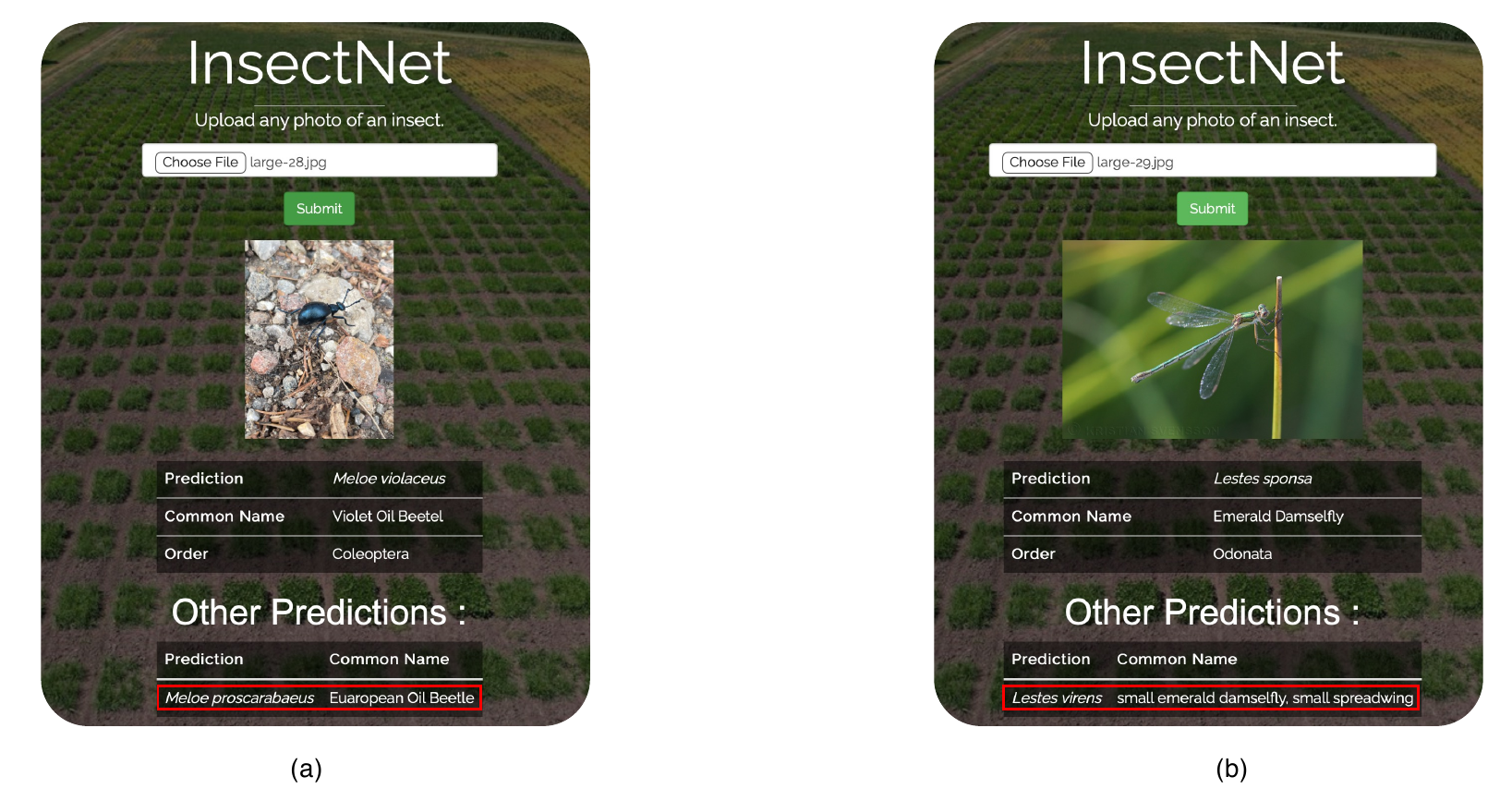}
  \caption{Instances of conformal prediction sets: Image of (a)\textit{Meloe violaceus} confused with \textit{Meloe proscarabaeus}, (b) \textit{Lestes sponsa} confused with \textit{Lestes virens}}
  \label{Fig:conformal_examples}
\end{figure*}

Our web app also has the functionality of conformal predictions integrated into it. When a sample is not detected to be OOD, it then returns a prediction set that contains all insect classes that have their conformal score lower than the threshold specified (\textit{\^{q}}). When an image of an ID insect class is uploaded, the model is highly certain of its prediction due to its exceptionally high MPC and returns only the true class label as shown in Fig \ref{Fig:conformal_examples}.a. In Fig \ref{Fig:conformal_examples}.b and \ref{Fig:conformal_examples}.c, samples of confounding insect classes that are often confused with other insect classes that are highly similar in appearance are depicted, hence the model is unsure of its prediction and returns the list of possible insect classes that the image could belong to.    

\section{InsectNet - an Insect-Pest Identification App}
We developed a web app that works on computers and handheld devices where users could use the insect class identifier. The interface is user-friendly with a minimalist design, wherein the user has to capture an image with an insect in it and submit it. The outcome of this submission is the prediction of an insect class and a prompt to let the user know that the submitted image could be an OOD sample if the model has low certainty on the sample, indicating that the submitted image could potentially have an image that the classifier has never seen before, for instance, an image of the plant, the face of a human or even an insect that was not in the training data. The prediction contains the scientific name and the common name for easier interpretability. Farmers could use this on a farm to get real-time predictions and when the internet connectivity is poor, the user could capture an image and later upload it to the server to get the prediction when they are connected to the internet. 
\begin{figure*}[ht!]
  \centering
  \includegraphics[width=10cm,height=6cm]{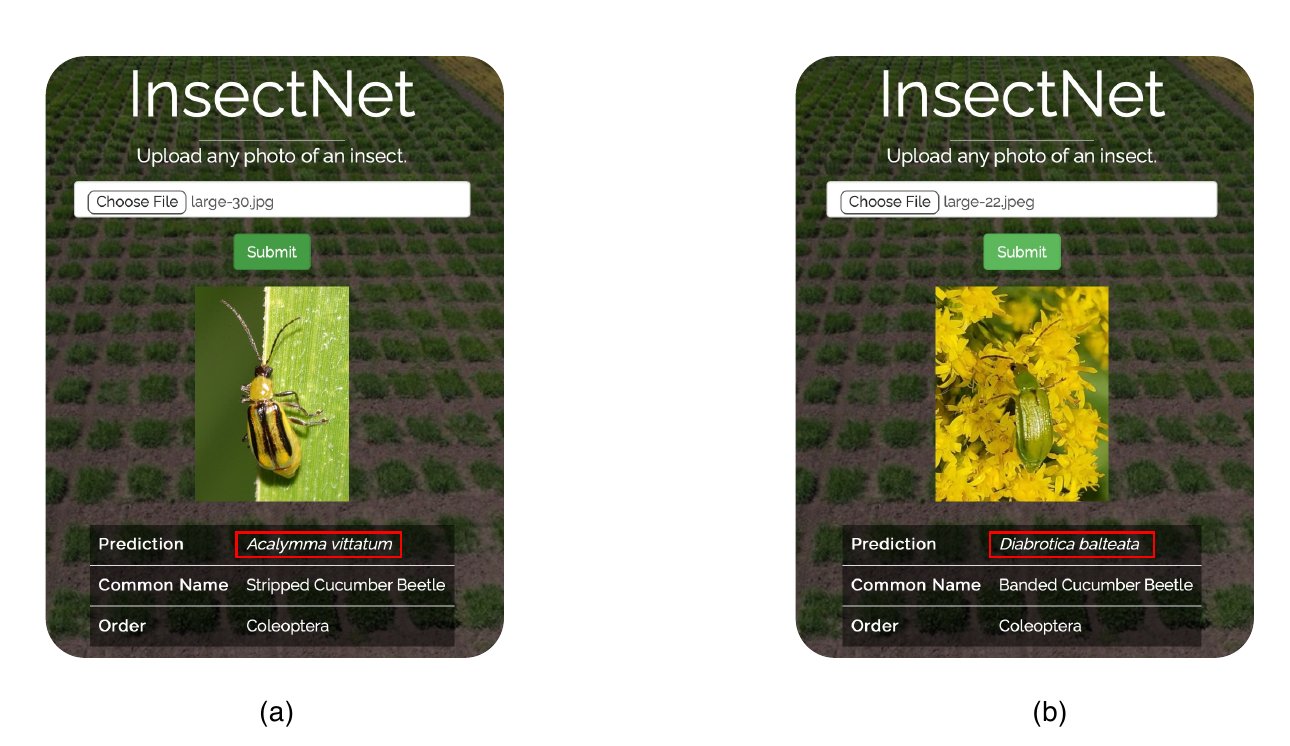}
  \caption{Instances of wrong predictions when InsectNet fails to identify (a) Northern corn root worm, (b) Western corn rootworm }
  \label{Fig:failure_examples}
\end{figure*}
\section{Examples of failure cases}
The InsectNet model struggles to differentiate between certain closely resembling insect pests that fall outside of its trained dataset and often provides confident but inaccurate predictions. For instance, while the model can identify the  \textit{Diabrotica balteata}(banded cucumber beetle) and \textit{Acalymma vittatum}(striped cucumber beetle), it lacks training on the  \textit{Diabrotica barberi}(Northern corn root worm) and \textit{Diabrotica virgifera}(Western corn rootworm ), which exhibit similar physical characteristics. It predicts northern corn root worm as banded cucumber beetle and western corn rootworm as striped cucumber beetle.

There are multiple ways to address these instances of model failures. The first is to extend the number of classes, which is a work in progress. The second is to incorporate additional filters -- in addition to OOD and conformal predictions -- that provide additional guardrails for prediction. We are currently exploring ensemble approaches.

\end{document}